\documentclass[letterpaper, 10 pt, conference]{ieeeconf}  

\IEEEoverridecommandlockouts                              

\usepackage{graphicx}
\graphicspath{ {./images/} }

\usepackage[]{algorithm2e}

\usepackage{multirow}
\usepackage{amssymb}
\usepackage{amsmath}
\usepackage{array}
\usepackage{mathtools}
\usepackage{subcaption}

\usepackage{tabularx}

\usepackage{xcolor,pifont}
\newcommand*\colourcheck[1]{%
  \expandafter\newcommand\csname #1check\endcsname{\textcolor{#1}{\ding{52}}}%
}
\newcommand*\colourcross[1]{%
  \expandafter\newcommand\csname #1cross\endcsname{\textcolor{#1}{\ding{55}}}%
}

\newcolumntype{C}[1]{>{\centering\arraybackslash}p{#1}}

\colourcheck{blue}
\colourcheck{black}
\colourcross{blue}
\colourcross{black}

\title{\LARGE \bf
DISCOMAN: Dataset of Indoor SCenes for Odometry, \\ Mapping And Navigation
}

\author{Pavel Kirsanov, Airat Gaskarov, Filipp Konokhov, Konstantin Sofiiuk, Anna Vorontsova, \\ Igor Slinko, Dmitry Zhukov, Sergey Bykov, Olga Barinova, Anton Konushin\\
\emph{Samsung AI Center}}

\begin{document}

\maketitle
\thispagestyle{empty}
\pagestyle{empty}

\begin{abstract}
We present a novel dataset for training and benchmarking semantic SLAM methods. The dataset consists of 200 long sequences, each one containing 3000-5000 data frames. We generate the sequences using realistic home layouts. For that we sample trajectories that simulate motions of a simple home robot, and then render the frames along the trajectories. Each data frame contains a) RGB images generated using physically-based rendering, b) simulated depth measurements, c) simulated IMU readings and d) ground truth occupancy grid of a house. Our dataset serves a wider range of purposes compared to existing datasets and is the first large-scale benchmark focused on the mapping component of SLAM. The dataset is split into train/validation/test parts sampled from different sets of virtual houses. We present benchmarking results for both classical geometry-based \cite{mur2017orb, engel2017direct} and recent learning-based \cite{costante2018ls} SLAM algorithms, a baseline mapping method \cite{zhou2018open3d}, semantic segmentation \cite{chen2018encoder} and panoptic segmentation \cite{sofiiuk2019adaptis}. The dataset and source code for reproducing our experiments will be publicly available at the time of publication. 
\end{abstract}

\IEEEpeerreviewmaketitle
\section{Introduction}

Simultaneous localization and mapping (SLAM) is an important component of robotic systems. Recently, the task of semantic SLAM has gained attention of the research community. It involves several components: trajectory estimation, mapping and semantic scene understanding. However, most of existing relevant datasets and benchmarks target distinct aspects of this complex task. Several benchmarks focus on trajectory estimation \cite{sturm12iros, geiger2013vision, handa2014benchmark, burri2016euroc, schubert2018vidataset}. The others target semantic understanding \cite{song2015sun, dai2017scannet, hernandez2017slanted}. Existing benchmarks for the mapping component of SLAM, e.g. Intel Lab Data \cite{intel2004data} are quite small and lack diversity. Evaluation of SLAM methods requires information about the poses of the camera. However, in order to obtain camera poses in indoor environments one needs special equipment, e.g. motion capture systems. For this reason real-world benchmarks for SLAM usually contain rather short trajectories across a small area (for instance, one room only). 

Recently computer graphics-generated datasets became popular for benchmarking computer vision models \cite{hernandez2017slanted}. It was shown that physically-based rendering can be successfully used for training computer vision models \cite{zhang2017physically}. Among the advantages of synthetic data are perfect ground truth annotation, control over difficulty and diversity of the data and an opportunity to obtain virtually unlimited number of samples. Over the last decade millions of designers have created an abundance of detailed and realistic 3d models of indoor environments. This wealth of data has a great potential for benchmarking semantic SLAM systems and improving the algorithms. 

\begin{figure}
    \centering
    \includegraphics[width=1.0\linewidth]{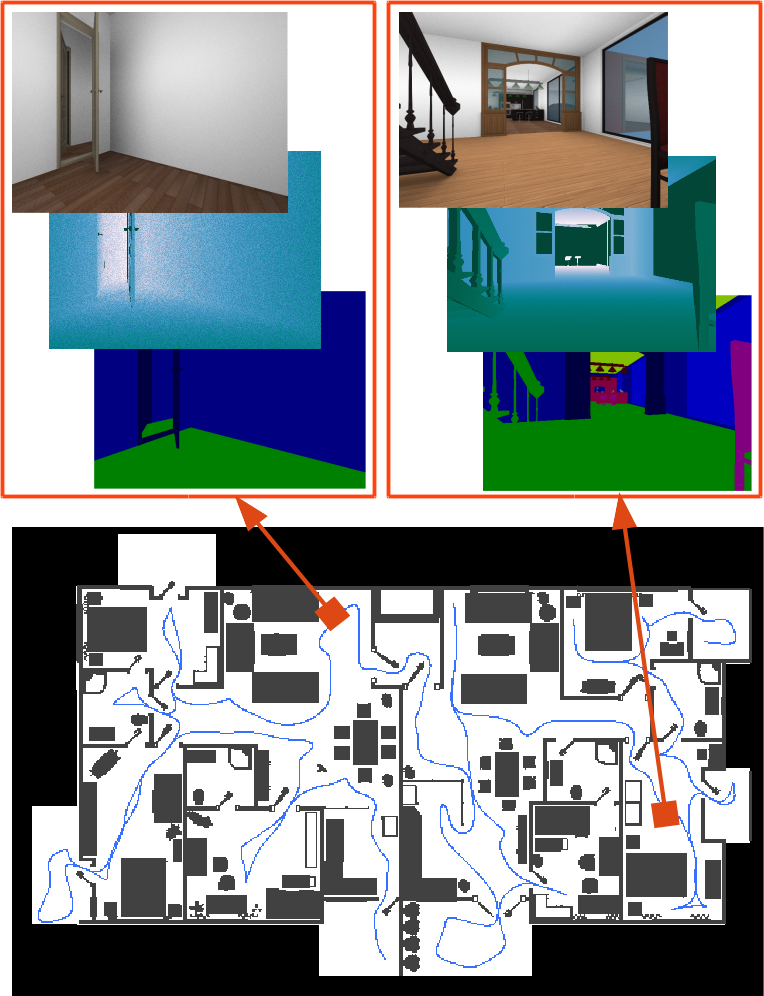}
    \caption{DISCOMAN dataset provides realistic indoor sequences with ground truth annotation for odometry, mapping and semantic segmentation.}
    \label{fig:discoman}
\end{figure}

In this work we present a new synthetic dataset called DISCOMAN (Dataset of Indoor SCenes for Odometry, Mapping And Navigation). It is generated using physically based image rendering with realistic lighting models. The data is obtained from the original home layouts created for refurbishment of real houses. We synthesize realistic trajectories as ground truth to render image sequences at video frame rate. In contrast to the existing datasets for SLAM that contain short sequences \cite{mccormac2017scenenet, li2018interiornet} we generate long trajectories simulating behaviour of a smart robot exploring a new home. The trajectories are more complex and diverse than in KITTI \cite{geiger2013vision}, but not as sophisticated as in hand-held datasets like TUM RGB-D \cite{sturm12iros} - see Figure \ref{fig:trajectories_kitti_discoman_tum}. Aside from rendering RGB images, we generate perfect and noised depth images and a pixel-accurate semantic annotation of object classes. We also generate ground truth occupancy grid for the visited part of a house. This can be used for training and benchmarking the mapping component of SLAM. Compared to existing benchmarks ours is an order of magnitude larger and much more diverse. It contains 200 long sequences, each of those contains about 3000-5000 data frames. This amount of data is enough for training and comprehensive evaluation of the models and at the same time is feasible to download and process. See Table \ref{tab:datasets_comparison} for comparison with existing datasets.

\begin{table*}[t]
  \centering
  \begin{tabular}{c|ccccC{0.7cm}C{0.8cm}C{0.6cm}C{0.6cm}C{0.6cm}C{0.8cm}C{0.7cm} }
    \hline
     & \textbf{Source} &  
     \begin{tabular}{@{}c@{}} \textbf{Camera} \\ \textbf{poses} \end{tabular} & 
     \begin{tabular}{@{}c@{}} \textbf{Motion} \\ \textbf{patterns} \end{tabular} &
     \begin{tabular}{@{}c@{}} \textbf{Frames} \\ \textbf{per sequence} \end{tabular} & 
     \begin{tabular}{@{}c@{}} \textbf{Large} \\ \textbf{scale} \end{tabular} &
     \begin{tabular}{@{}c@{}} \textbf{Scene} \\ \textbf {diversity} \end{tabular} &
     \textbf{Depth} & 
     \textbf{Stereo} &
     \textbf{IMU} & 
     \begin{tabular}{@{}c@{}} \textbf{Semantic} \\ \textbf{labelling} \end{tabular} &
     \begin{tabular}{@{}c@{}} \textbf{2d map} \end{tabular} \\
    \hline
    \textbf{TUM RGB-D} \cite{sturm12iros} & 
            real & 
            \begin{tabular}{@{\vspace{0em}}c@{}} motion \\ capture \end{tabular} & 
            \begin{tabular}{@{\vspace{0em}}c@{}} hand-held, \\ robot\end{tabular}&
            $\sim$1000 &
            &
            &
            \blackcheck &
            &
            &
            &
            \\
    [\bigskipamount]
    \textbf{TUM VI} \cite{schubert2018vidataset} &
            real &
            \begin{tabular}{@{\vspace{0em}}c@{}} motion \\ capture \end{tabular} & 
            hand-held &
            $\sim$2000 &
            &
            &
            &
            \blackcheck &
            \blackcheck &
            &
            \\
    [\bigskipamount]
    \textbf{EuRoC} \cite{burri2016euroc} & 
            real & 
            \begin{tabular}{@{\vspace{0em}}c@{}} motion \\ capture \end{tabular} & 
            MAV & 
            $\sim$3000 &
            &
            &
            &
            \blackcheck &
            \blackcheck &
            &
            \\
    [\bigskipamount]
    \textbf{ScanNet} \cite{dai2017scannet} & 
            real & 
            \begin{tabular}{@{\vspace{0em}}c@{}} structure \\ from motion \end{tabular} &
             \begin{tabular}{@{\vspace{0em}}c@{}} hand-held \\ 3d scanning \end{tabular} &
            $\sim$1500 &
            \blackcheck &
            \blackcheck &
            \blackcheck &
            &
            &
            \blackcheck &
            \\
    [\bigskipamount]
    \textbf{ICL-NUIM} \cite{handa2014benchmark} &
            render &
            ground truth &
            random &
            $\sim$1000 &
            &
            &
            \blackcheck &
            &
            &
            &
            \\
    [\bigskipamount]
    \textbf{SceneNet RGB-D} \cite{mccormac2017scenenet} & 
            render & 
            ground truth &
            random & 
            300 &
            \blackcheck  &
            &
            \blackcheck &
            &
            &
            \blackcheck  &
            \\
    [\bigskipamount]
    \textbf{InteriorNet} \cite{li2018interiornet} & 
            render & 
            ground truth &
            random &
            1000 & 
            \blackcheck &
            \blackcheck &
            \blackcheck &
            \blackcheck &
            \blackcheck &
            \blackcheck &
            \\
    [\bigskipamount]
    \hline
    \textbf{DISCOMAN} & 
            render &
            ground truth &
            \textbf{robot} & 
            \textbf{3000-5000} &
            \blackcheck &
            \blackcheck &
            \blackcheck &
            \blackcheck &
            \blackcheck &
            \blackcheck &
            \blackcheck \\
    \hline
  \end{tabular}
  \caption{Comparison of indoor datasets with camera poses.}
  \label{tab:datasets_comparison}
\end{table*}

\begingroup
\setlength{\tabcolsep}{0pt} 
\renewcommand{\arraystretch}{1.5} 
\begin{figure}
  \begin{tabular}{cc}
    \includegraphics[width=0.48\linewidth]{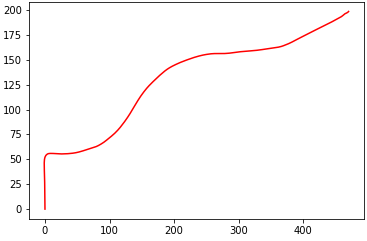} &
    \includegraphics[width=0.48\linewidth]{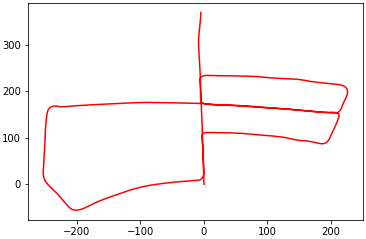} \\
    \includegraphics[width=0.48\linewidth]{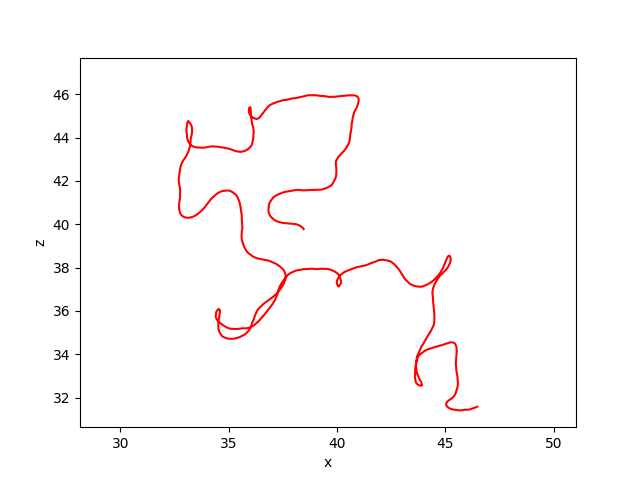} &
    \includegraphics[width=0.48\linewidth]{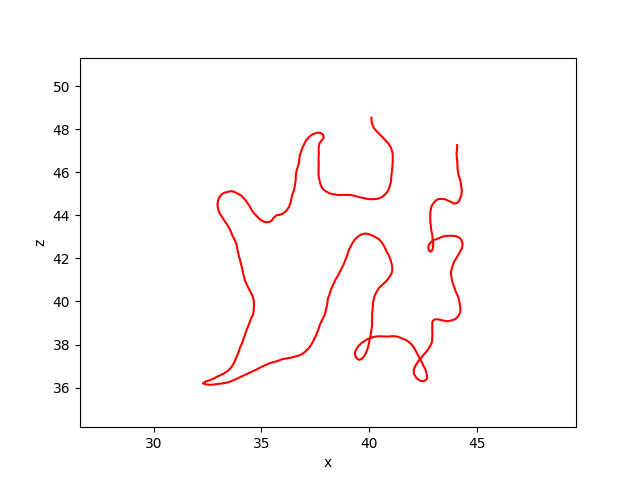} \\
    \includegraphics[width=0.48\linewidth]{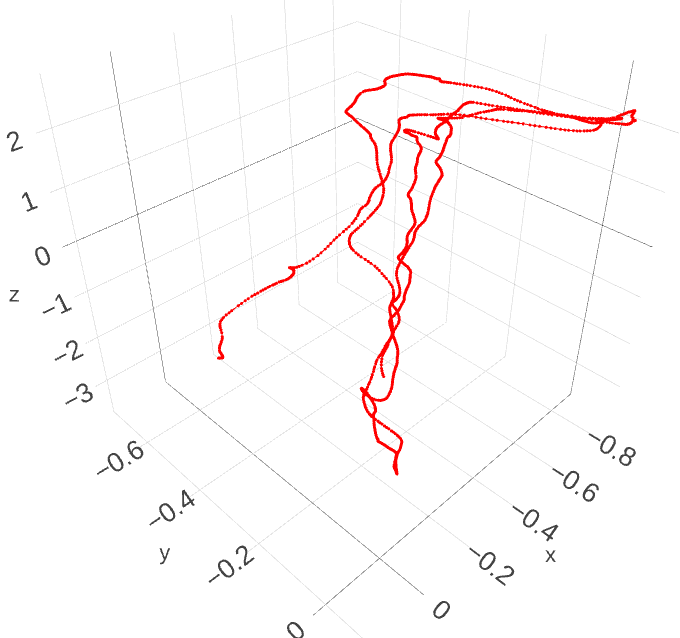} &
    \includegraphics[width=0.48\linewidth]{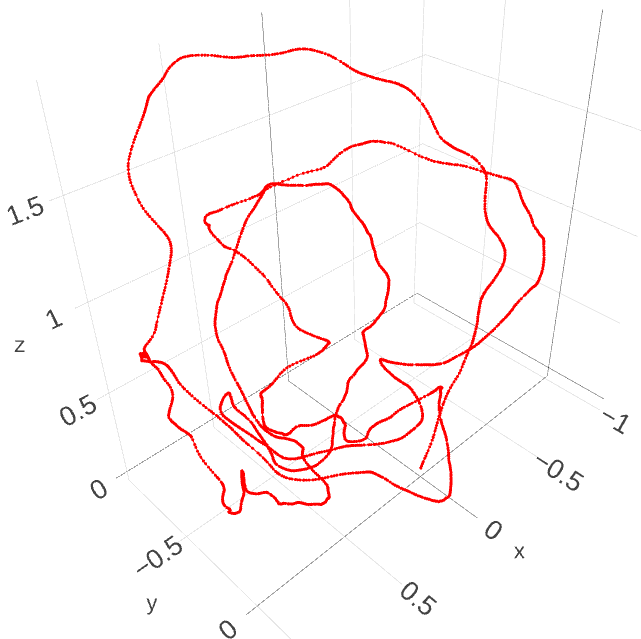} \\
  \end{tabular}
  \caption{Sample trajectories from outdoor KITTI \cite{geiger2013vision} (top row), and indoor DISCOMAN (middle row) and TUM RGB-D \cite{sturm12iros} (bottom row) benchmarks. The trajectories in DISCOMAN are slightly more difficult compared to KITTI, but less complex compared to TUM RGB-D.}
  \label{fig:trajectories_kitti_discoman_tum}
\end{figure}
\endgroup

Multiple algorithms for constructing maps have been proposed \cite{de2017skimap,maier2012real,hornung2013octomap}, some of them are based on deep learning \cite{lu2018monocular, erkent2018semantic, zhang2018egocentric}. However benchmarking of these methods is currently complicated due to lack of a suitable dataset. Since there are no conventional metrics for evaluating an accuracy of mapping algorithms, in this work we introduce and describe the new set of metrics for evaluation of mapping accuracy. We believe that our benchmark can bring new insights and facilitate development of more accurate and robust methods for mapping.

Using the generated dataset we perform a comprehensive evaluation of current state-of-the-art methods. Our evaluation includes visual SLAM/odometry methods, namely classical ORBSLAM \cite{mur2017orb} and more recent learning-based method \cite{costante2018ls}, an Open3D-based method for mapping \cite{zhou2018open3d} and a state-of-the-art semantic segmentation method \cite{chen2018encoder}. These results can be used as a baseline for further research.

The rest of the paper is organized as follows. In section \ref{related} we discuss related works. In section \ref{description} we describe in details the process of data generation that involves trajectories sampling and rendering. Section \ref{experiments} is devoted to experiments, and section \ref{conclusion} is left for conclusions.

\section{Related work} \label{related}

The closest work to ours is InteriorNet \cite{li2018interiornet}, that presents a mega-scale indoor dataset containing a large number of short synthetic sequences. Similarly to our work they used physically based rendering for data synthesis. However, it is worth noting that only a small number of InteriorNet sequences are now available for public use and all of them are based on randomized motions. Compared to InteriorNet we focus more on the mapping component of SLAM. Thus, we generate longer sequences (about 3000-5000 frames length compared to 1000 frames in InteriorNet) and provide ground truth maps along with the sequences.

\begingroup
\setlength{\tabcolsep}{2pt} 
\renewcommand{\arraystretch}{1.5} 
\begin{figure*}[t]
\begin{tabular}{c c c}
  \includegraphics[width=0.325\linewidth]{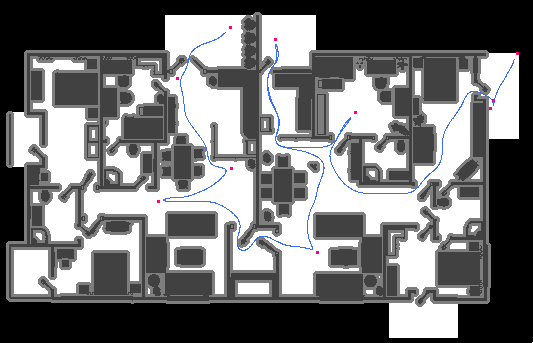} &
  \includegraphics[width=0.325\linewidth]{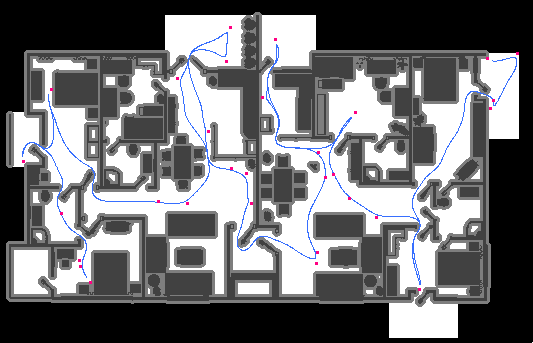} &
  \includegraphics[width=0.325\linewidth]{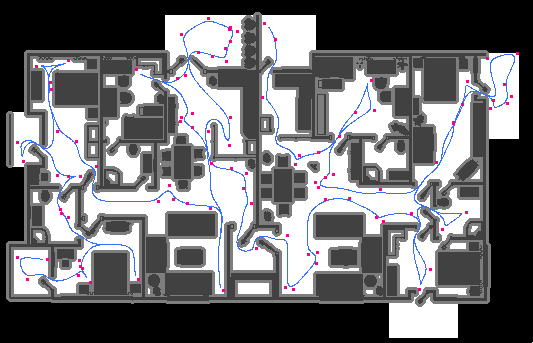} \\
  (a) & (b)  & (c)
  \end{tabular}
  \caption{Samples of generated trajectories. Color coding:  red - sampled keypoints, blue - final trajectory after smoothing, black - occupied areas, white - free area, grey - the area of an image where keypoints cannot be sampled. One can see the effect of choosing different number of keypoints per trajectory: (a) 10 keypoints, (b) 30 keypoints, (c) 100 keypoints per trajectory. One can see that the more keypoints we add, the more curved the trajectory gets.}
  \label{fig:sampling}
\end{figure*}
\endgroup

Another great example of a multi-purpose dataset is the renowned KITTI benchmark suite \cite{geiger2013vision}. It provides real data with different types of annotation including camera poses for evaluation of SLAM/odometry methods, semantic/panoptic segmentation and object bounding boxes in 2d and 3d. However this dataset is highly specialized for self-driving, e.g. the trajectories are composed mainly of straight lines and contain very few turnings. Another problem with KITTI is low diversity of the sequences. It contains only 22 sequences taken in very similar conditions. In this work we provide an order of magnitude more sequences sampled from diverse indoor environments.

The most popular real-world indoor datasets for evaluation of trajectory estimation are
the TUM RGB-D benchmark \cite{sturm12iros} containing RGB-D sequences and EuRoC \cite{burri2016euroc} containing stereo+IMU sequences. TUM RGB-D contains both hand-held trajectories and the trajectories taken from a robotic platform. The recent TUM VI \cite{schubert2018vidataset} dataset is desinged for benchmarking visual intertial odometry. The popular synthetic ICL-NUIM dataset \cite{handa2014benchmark} has a few RGB-D sequences with modelled noise of a depth sensor. All sequences in ICL-NUIM are sampled from randomized trajectories across two 3d models. Each of the mentioned real-world datasets contain about a dozen sequences. The small scale of the datasets and lack of diversity makes it difficult to reason about the robustness of SLAM methods. 

ScanNet dataset \cite{dai2017scannet} is a great effort to collect 3d models of real houses using structure-from-motion technique. This benchmark targets semantic and instance segmentation in 2d and 3d. But the trajectories in this dataset are highly specific for 3d scanning applications with abundance of loopy motion patterns that are not relevant to robotic applications. Compared to ScanNet we focus more on robotic applications and create trajectories accordingly. Our dataset contains longer sequences with more robot-like motion patterns.

Matterport3D \cite{Matterport3D} and 2D-3D-S \cite{2017arXiv170201105A} are the other examples of datasets collected with a 3d scanner, i.e. Matterport camera. They provide 3D real-world scenes with raw 3D point clouds, segmentation and reconstructed meshes. However the 3d scanning process with Matterport cameras does not produce smooth trajectories, and the quality of 3d models does not allow for interpolation between the frames.

A few synthetic datasets relevant to this work have been proposed. SceneNet RGB-D \cite{mccormac2017scenenet} contains millions of frames organized in sequences corresponding to complex camera trajectories. This dataset was generate using randomly cluttered furniture, thus the main drawback of SceneNet RGB-D is low realism. Virtual KITTI \cite{gaidon2016virtual} is a synthetic dataset of outdoor scenes labeled with accurate  ground  truth for  object detection, tracking, scene and instance segmentation, depth and optical flow. It is also worth to mention, that a number of simulators for reinforcement learning have appeared recently.

\begingroup
\setlength{\tabcolsep}{1pt} 
\setlength\extrarowheight{-6pt}
\begin{figure*}
  \begin{tabular}{cccccc}
  \includegraphics[width=0.195\linewidth]{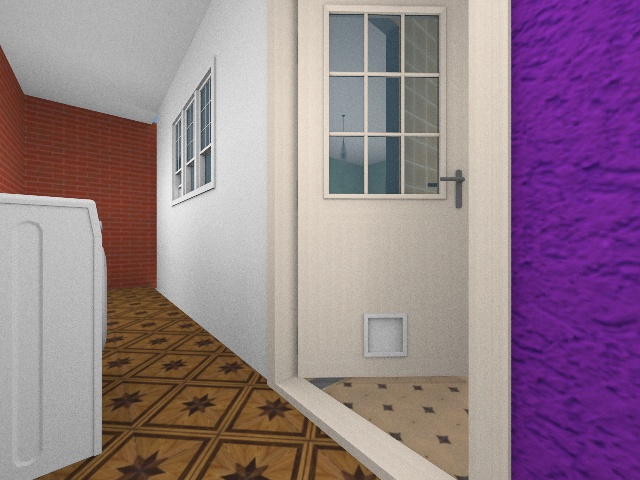} &
    \includegraphics[width=0.195\linewidth]{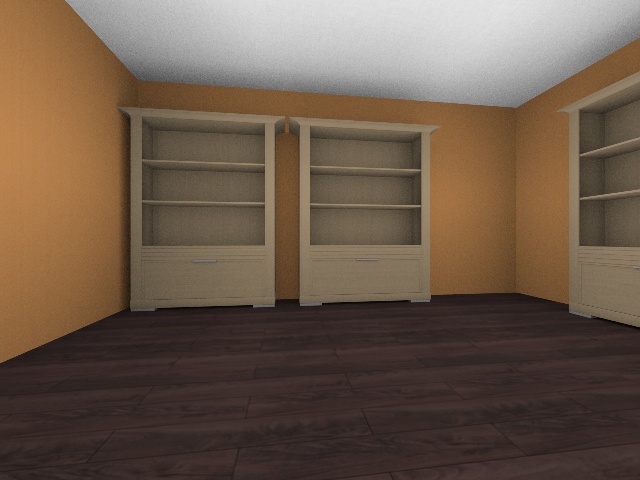} &
   \includegraphics[width=0.195\linewidth]{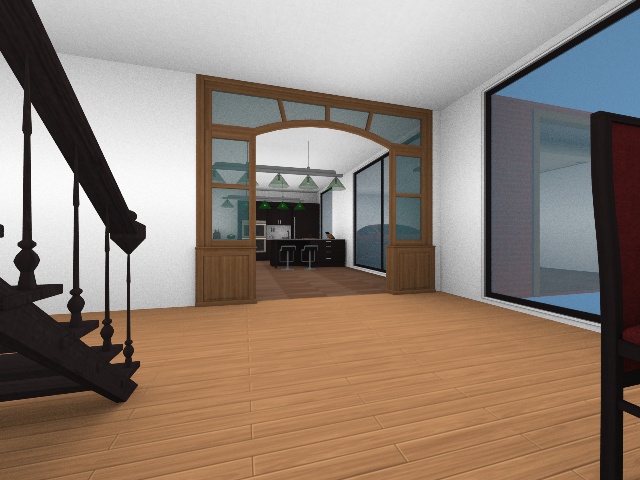} &
    \includegraphics[width=0.195\linewidth]{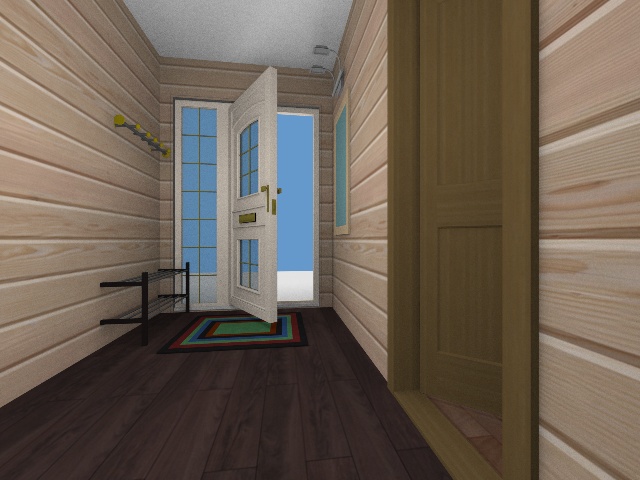} &
   \includegraphics[width=0.195\linewidth]{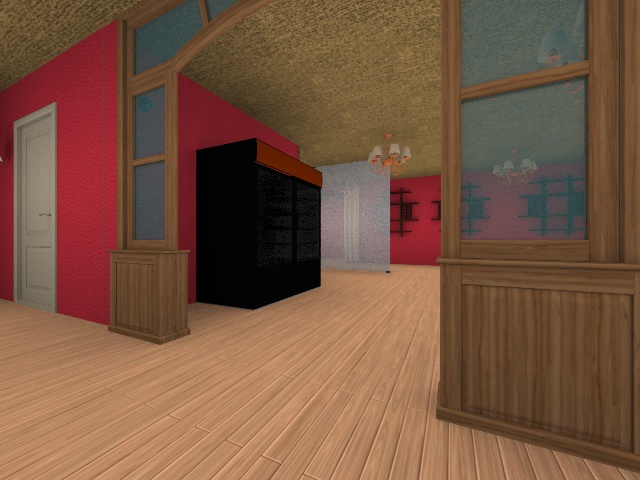} \\
  \includegraphics[trim={0.6cm 0.6cm 0 0},clip, width=0.195\linewidth]{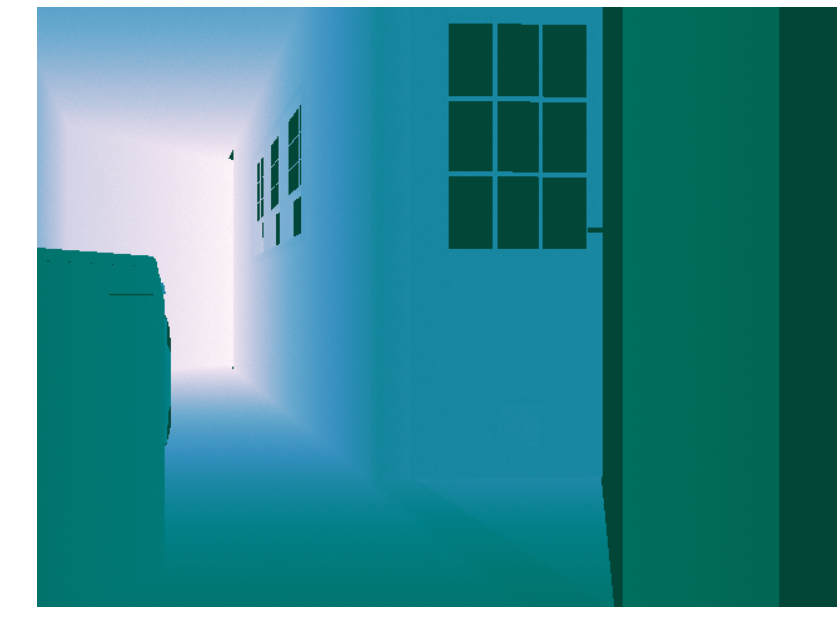} &
  \includegraphics[trim={0.6cm 0.6cm 0 0},clip, width=0.195\linewidth]{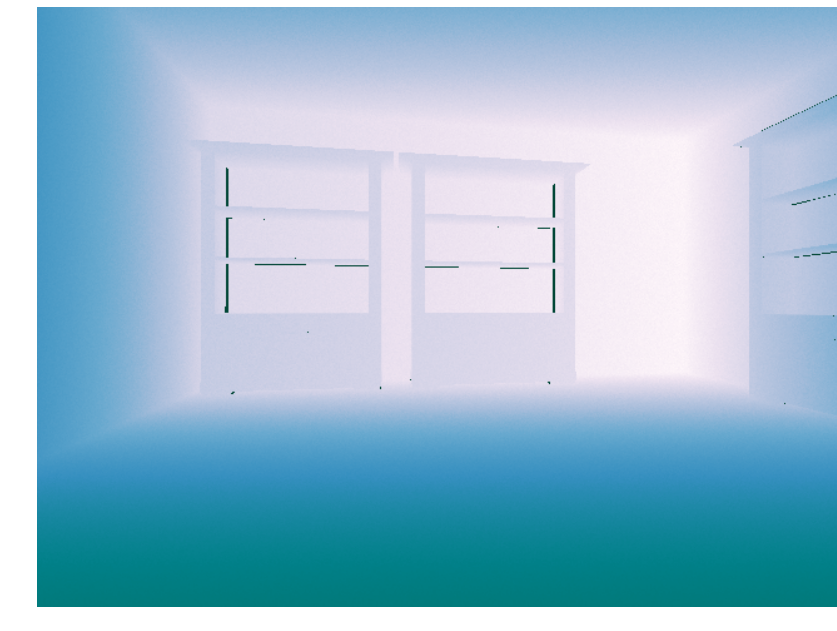} &
  \includegraphics[trim={0.6cm 0.6cm 0 0},clip, width=0.195\linewidth]{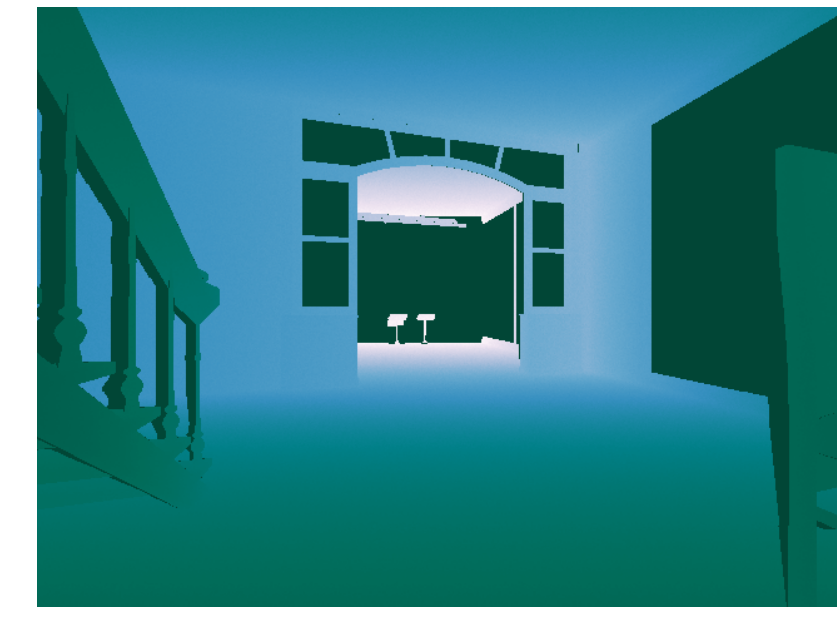} &
  \includegraphics[trim={0.6cm 0.6cm 0 0},clip, width=0.195\linewidth]{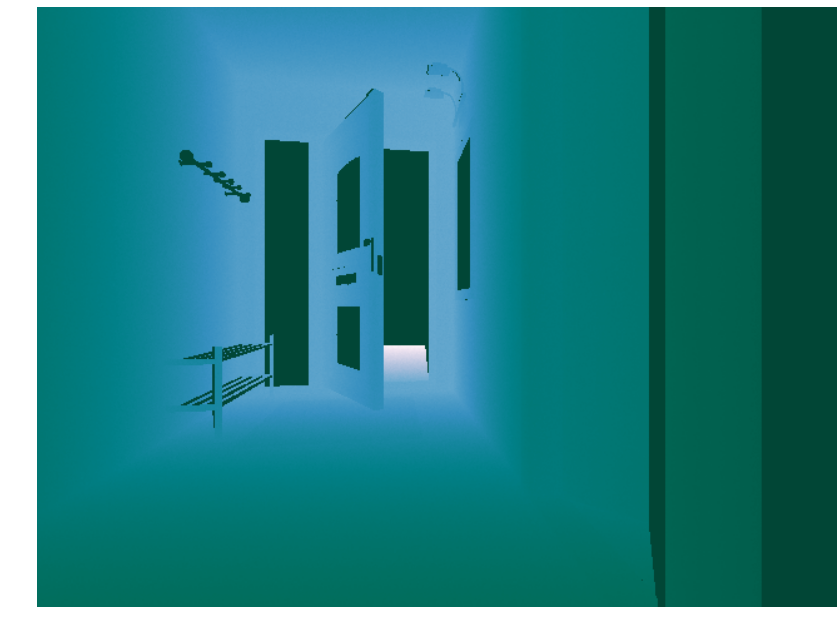} &
  \includegraphics[trim={0.6cm 0.6cm 0 0},clip, width=0.195\linewidth]{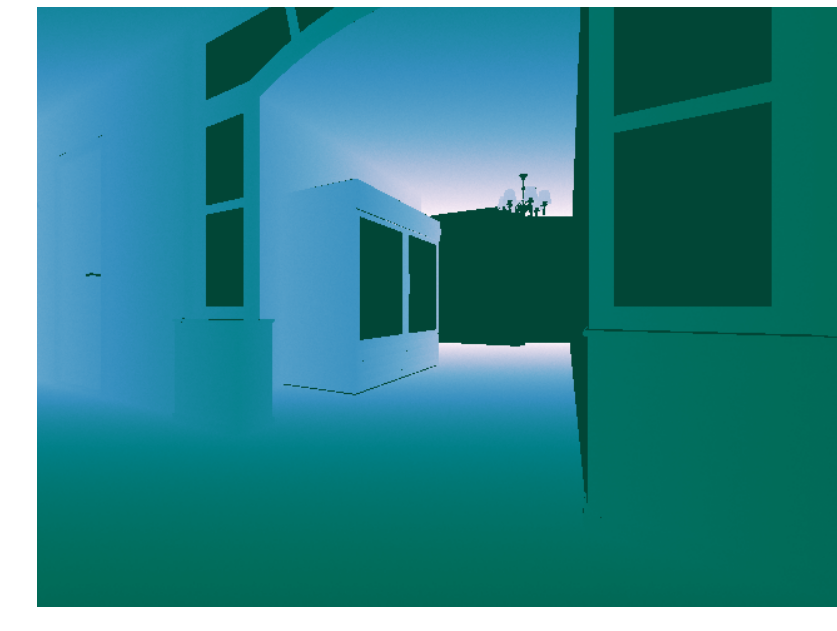}\\
  \includegraphics[width=0.195\linewidth]{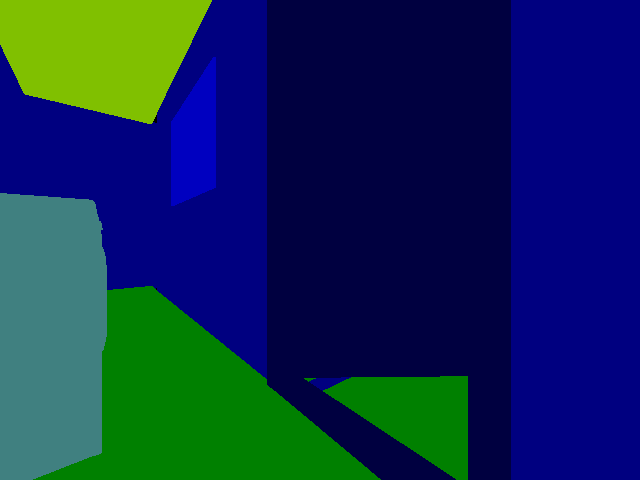} &
  \includegraphics[width=0.195\linewidth]{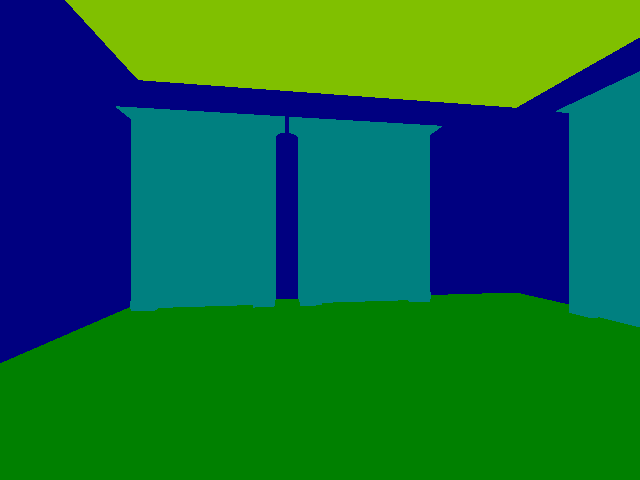} &
  \includegraphics[width=0.195\linewidth]{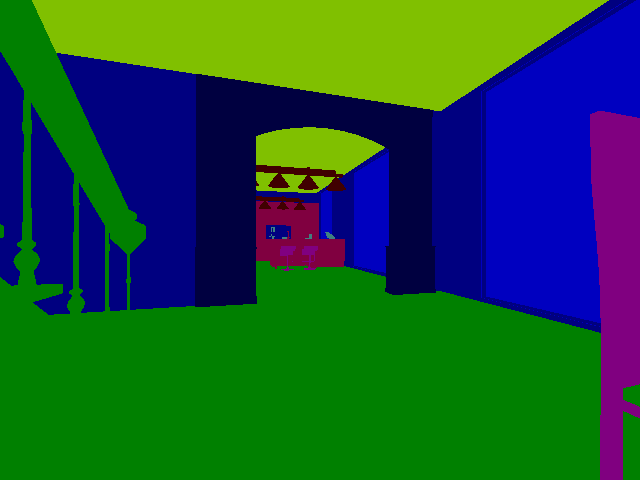} &
  \includegraphics[width=0.195\linewidth]{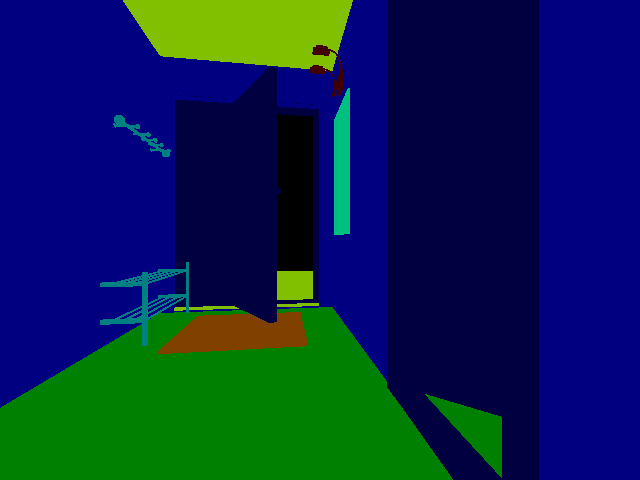} &
  \includegraphics[width=0.195\linewidth]{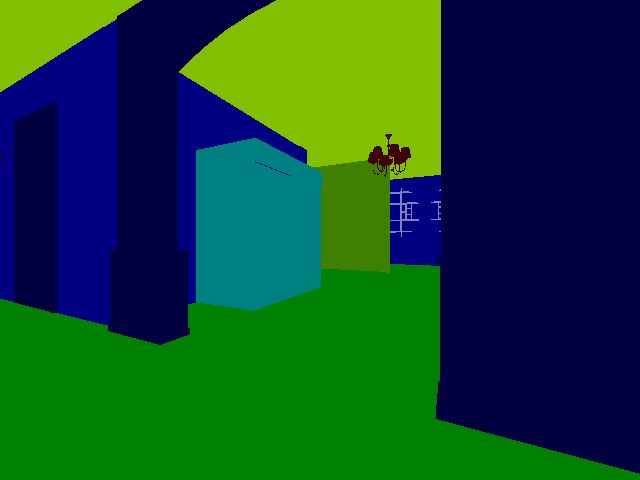}
  \end{tabular}
  \caption{Example frames from DISCOMAN dataset. From top to bottom: RGB image, depth with emulated sensor noise, pixel-wise semantic annotation. Notice holes in depth maps for reflecting and black surfaces.}
  \label{fig:frames}
\end{figure*}
\endgroup

\section{Dataset generation} \label{description}

\textbf{Trajectories generation algorithm.}
Our goal is to realistically model motions of a robot within a given scene. An algorithm that we use for trajectories generation includes the following steps. First, we compute 3D occupancy grid within scene bounding box with constant size of a grid cell (5cm in our work). Then we find traversable grid cells, i.e the ones that lie not closer than a given distance to the obstacles (20cm in our work). Next, we uniformly sample $N$ random points from a set of traversable nodes. Point count depends on scene accessible area. We have noticed that such point density (point count per square meter) is highly correlated with trajectory complexity, i.e. linear and angular acceleration/deceleration. Then we apply travelling salesman problem (TSP) solver algorithm to find the order for visiting points, so each point is visited only once. After that we compute weighted shortest path passing through sampled points. The weights are inversely proportional to the distance between the agent and the closest obstacle. Finally we generate path between the points using full state planning algorithm, which takes into account linear/angular velocity/acceleration limits for a given robot. Each trajectory can be sampled with desired time resolution between frames. We choose 150 Hz sampling rate  for IMU data representation and 30 Hz for image sensor data representation.

\textbf{Rendering.}
We have developed a custom visualization engine named Renderbox. It is capable of producing various robotics-specific data as well as generating true physically-based shaded images. Renderbox consists of two image generation back-ends: multi-threaded CPU raytracing renderer adapted for cluster infrastructure and a GPU accelerated rasterization renderer. Both of them use the same scene graph, which made possible smooth and instantaneous data transitions through the whole rendering pipeline. 

RGB images are generated using raytracing algorithm. We chose bidirectional path-tracing with pre-gathered and pre-filtered photon maps as a good compromise between suitable performance rate and visually pleasing results. For solving ray-triangle intersection problem we use Intel Embree library. Our physically-based rendering model allows us to vary scene visual representation conditions by applying a number of effects. Currently, it uses approximation of ambient occlusion effect which provides realistically looking images. While the raytracing back-end is used for rendering RGB images, depth and segmentation maps are generated in real-time using OpenGL API. Examples of rendered data are shown in Figure \ref{fig:frames}.

\section{Dataset description}

The dataset is split into train, validation and test parts and is designed for the following tasks: trajectory estimation, mapping and semantic segmentation. 

\subsection{Trajectory estimation}

We formulate this task as follows. Given an input sequence one needs to estimate corresponding positions and orientations of a robot.

\textbf{Metrics.} To compute the metrics, the estimated and ground truth trajectories  first  need to be aligned. We use Horn method \cite{horn1987closed}, which  finds  the  rigid-body  transformation $S$. Then we compute standard ATE (Absolute Trajectory Error) and RPE (Relative Pose Error) metrics. Below we formally define those metrics to avoid ambiguity.

Let us define absolute trajectory error matrix at time $i$ as:
\begin{equation*}
E_i \coloneqq Q_i^{-1}SP_i
\end{equation*}
The ATE is defined as the root mean square error from error matrices:
\begin{equation*}
\mbox{ATE}_{rmse} = \bigg(\frac{1}{n} \sum_{i=1}^{n} \parallel \mbox{trans}(E_i) \parallel^2 \bigg)^{\frac{1}{2}}
\end{equation*}
Actually, absolute trajectory error is the average deviation from ground truth trajectory per frame.

The relative pose error measures the local accuracy of the trajectory over a fixed time interval $\Delta$. Therefore, the relative pose error corresponds to the drift of the trajectory which is in  particular  useful  for  the  evaluation  of  visual  odometry systems.  Let us define the relative pose error matrix at time step $i$ as:
\begin{equation*}
F_i^{\Delta} \coloneqq (Q_i^{-1} Q_{i+\Delta})^{-1}(P_i^{-1}P_{i+\Delta})
\end{equation*}
from  a  sequence  of $n$ camera  poses we obtain $m = n - \Delta $ individual  relative  pose  error matrices  along  the sequence. The RPE is usually divided into translation and rotation components. Similar  to  the  absolute trajectory  error,  we  propose  to  evaluate the  root  mean  squared  error  over  all  time  indicies for RPE translation error:
\begin{equation*}
\mbox{RPE}_{trans}^{i, \Delta} = \bigg(\frac{1}{m} \sum_{i=1}^{m} \parallel \mbox{trans}(F_i) \parallel^2 \bigg)^{\frac{1}{2}}
\end{equation*}
As for rotation component we use mean error approach:
\begin{equation*}
\mbox{RPE}_{rot}^{i, \Delta} = \frac{1}{m} \sum_{i=1}^{m} \angle (\mbox{rot}(F_i^{\Delta}))
\end{equation*}
We average over all possible pairs in both translation and rotation component.

\subsection{Mapping}
In this work we focus on estimation of 2D occupancy maps as they are commonly used for motion planning and navigation. We consider maps with two states of cells: ``empty'' and ``occupied''. The task is formulated as follows. Given a sequences of inaccurate camera poses and raw RGB-D frames with noisy depth, the goal is to reconstruct 2D occupancy map of the visited part of a indoor scene.

\textbf{Map scaling and alignment.} To evaluate the quality of mapping result we need to align, scale and offset the predicted map with ground truth map. This task raises a challenge to compare both maps in the same coordinate system with appropriate scales and offsets. 

Using the agent initial position and orientation in world coordinate system provided in the DISCOMAN dataset we transform all the frames from camera coordinate system to world coordinate system. Such transformation ensures to have all the point cloud extracted from frames are in the same coordinate system with ground truth.

To evaluate mapping results with ground truth map we further need to apply transformation to grid coordinate system. Then we project the point cloud to 2D coordinate system representing the predicted map. We provide transformation from world coordinate system to grid coordinate system as 4x4 matrix in the GRD file as part of DISCOMAN dataset. Also this matrix contains appropriate scale and offset for matching and centering a predicted map to ground truth map.  

The described sequence of transformation ensures that the resulting map is aligned with the ground truth map. This removes potential artifacts that could arise from manipulating with images in 2D space such as map rotations and welding. This simplifies map quality evaluation and makes it more accurate.

\textbf{Metrics.}
For evaluation of mapping results we use a modified version of Map Score metric introduced in \cite{colleens2007occupancy}. Map score gives a positive value representing the difference between two maps (generally the ground truth map of the environment and the generated map that we are evaluating), so the lower the number, the more alike the two maps are. To normalise the score, we compute the worst possible map that could be compared to the ground truth map among the three variants: a map with inverted values of occupancy grid in the dilated occupied regions \cite{colleens2007occupancy}, an empty map, and fully occupied map. The value of Map Score for the evaluated map is then divided by the maximum of the Map Scores for these three maps.

\subsection{Semantic/panoptic segmentation}

We formulate the task as follows. Given an input sequence of frames one needs to predict the pixel-wise semantic/panoptic segmentation labelling for each frame. We perform evaluation across sequences, therefore the previous frames in the sequence can be utilized to achieve higher accuracy. 

\textbf{Ground truth annotation.} Each RGB image in DISCOMAN comes along with corresponding pixel-wise semantic annotation. We used an ontology very similar to the one suggested in NYUv2 dataset \cite{Silberman:ECCV12}. The dataset provides annotation for both semantic and instance segmentation tasks. We also split the classes into `things` and `staff` and provide annotation for panoptic segmentation \cite{kirillov2019panoptic}.

\textbf{Metrics.} In order to provide more diverse data for training semantic segmentation models we have generated additional dataset consisting of 60000 images taken from 12000 different scenes. For testing we use every 10th frame in the test sequences. For semantic segmentation we compute standard metrics, i.e. mIoU and pixel accuracy. For panoptic segmentation we compute PQ, SQ and RQ metrics as suggested in \cite{kirillov2019panoptic}.

\section{Experiments} \label{experiments}

\subsection{Trajectory estimation}

\begin{table*}
\begin{center}
\begin{tabular}{l|cccc}
Method   &   Success rate  &   ATE  &   RPE-t  &  RPE-r (deg) \\
    \hline
    \hline
DSO (mono)  &   72\%  &  2.59  &  8.11  &  93.84  \\
LS-VO (mono) &   100\%  &  1.11  &  1.67 &  18.43  \\
ORBSLAM2 (RGB-D)  &   11\%  &  0.69  &  1.13  &  11.20  \\
Motion Maps (RGB-D) &   100\%  &  0.82  &  1.17  &  14.25  \\
    \hline
Motion Maps - ORBSLAM2 sequences &   100\%  &  0.32  &  0.48  &  4.19  \\
Motion Maps - DSO sequences &   100\%  &  0.42  &  0.52  &  5.9  \\
    \hline
\end{tabular}
\end{center}
\caption{Comparison of ORBSLAM2 \cite{mur2017orb}, DSO \cite{engel2017direct}, LS-VO \cite{costante2018ls} and Motion Maps \cite{slinko2019scene} methods. We compute ATE, RPE rotation and RPE translation. DSO and ORBSLAM2 fail due to tracking loss on several sequences, therefore we report success rate for every method. We additionally report accuracy of Motion Maps method for the sequences where DSO or ORBSLAM2 succeeded.}
\label{tab:trajectory_results}
\end{table*}

\textbf{Evaluated methods.}
We compute results for DSO (monocular) \cite{engel2017direct}, ORBSLAM2 (RGB-D) \cite{mur2017orb} and recent learning-based LS-VO (monocular) \cite{costante2018ls} and Motion Maps  (RGB-D) \cite{slinko2019scene} method. We used author implementations for both evaluated methods in our experiments. In our experiments we used PWC-Net \cite{Sun2018PWC-Net} for optical flow estimation in LS-VO and Motion Maps.

\begingroup
\setlength{\tabcolsep}{2.5pt} 
\begin{figure*}[t]
  \begin{tabular}{ccc}
  \includegraphics[width=0.325\linewidth]{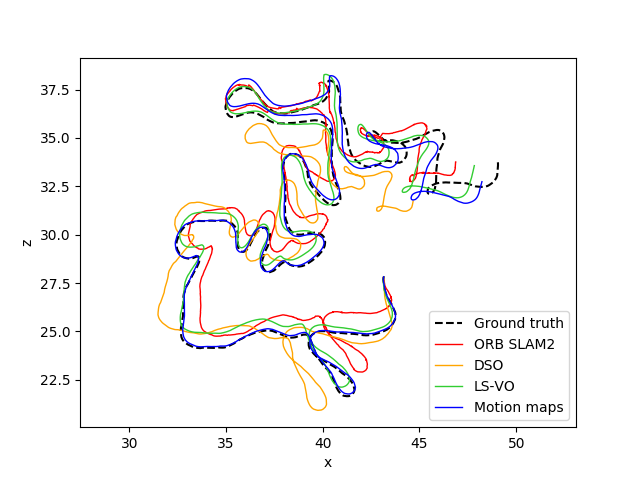} &
  \includegraphics[width=0.325\linewidth]{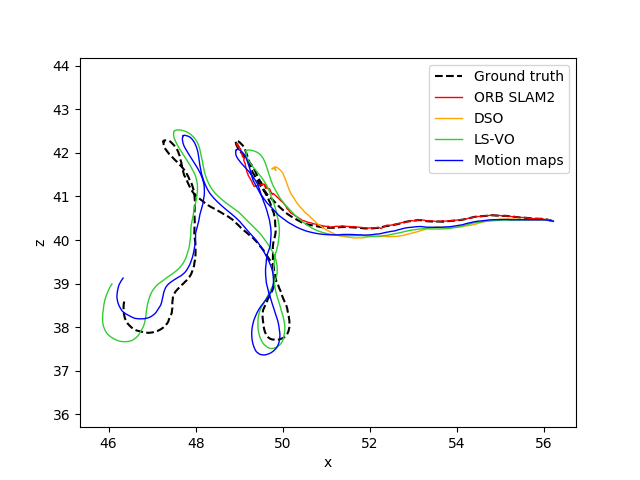} &
  \includegraphics[width=0.325\linewidth]{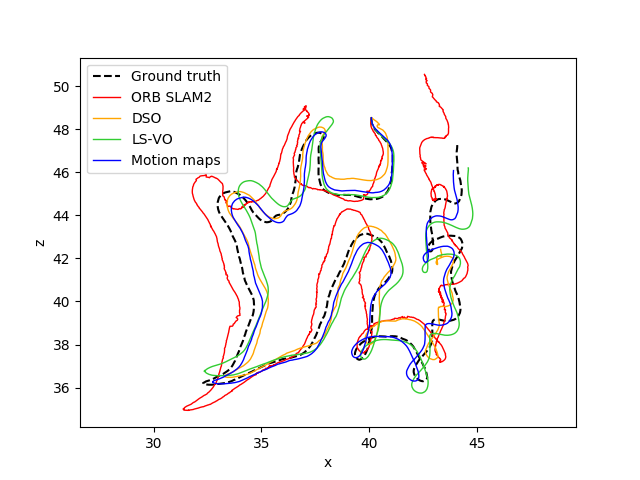} \\
  (a) & (b) & (c) \\
  \includegraphics[width=0.325\linewidth]{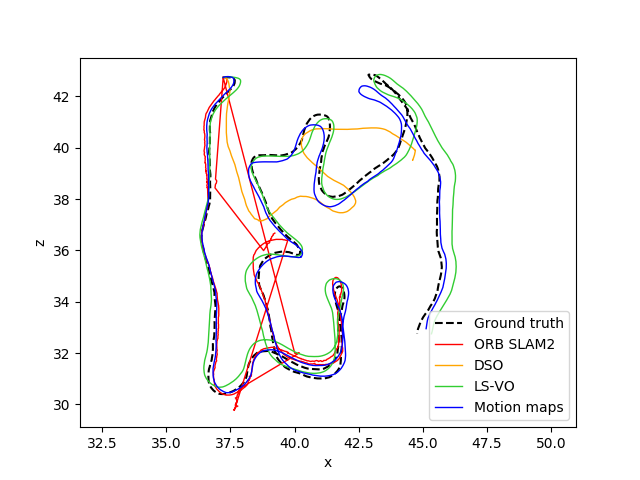} &
  \includegraphics[width=0.325\linewidth]{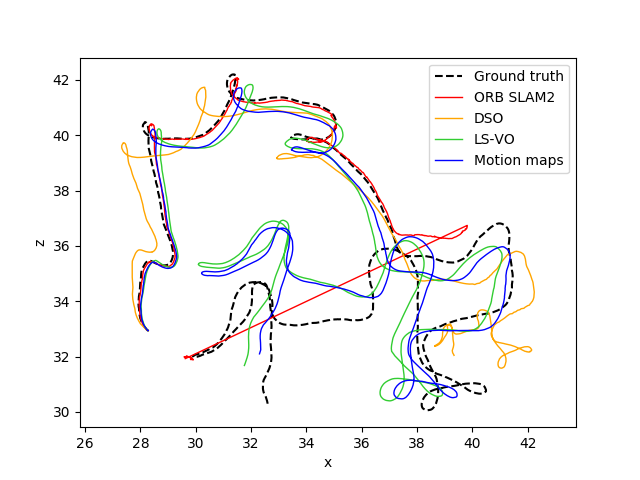} &
  \includegraphics[width=0.325\linewidth]{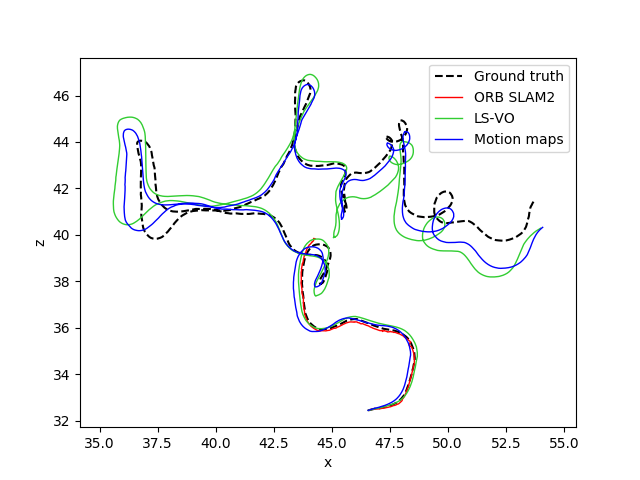} \\
  (d) & (e) & (f)
  \end{tabular}
  \caption{Qualitative results of trajectory estimation. One can see that DISCOMAN dataset is difficult for sparse SLAM methods like DSO (monocular) and ORBSLAM2 (RGB-D). The main reasons for that are the abundance of fast rotations and low-textured surfaces, e.g. white walls. Learning-based methods LS-VO (monocular) and Motion Maps (RGB-D) show higher robustness, but in most cases lower accuracy.}
  \label{fig:trajectory_images}
\end{figure*}
\endgroup

\textbf{Details and results.}
Since ORBSLAM2 is randomized, each test sequence in the dataset is processed 10 times to find the median value for each metric. 
We trained LS-VO and Motion Maps on the train part of the data with initial learning rate = $0.001$ using Adam with default parameters (beta1 = $0.9$, beta2 = $0.99$). We used two separate L2 losses for translation and rotation components (Euler angles) of the motion with rotation loss multiplied by 50. The following LR scheduling was used: learning rate was multipled by 0.5 if validation loss does not decrease for 10 epochs. We have trained the model for 100 epochs on 1 GPU with batch size $128$.

Results of the evaluation are shown in Table \ref{tab:trajectory_results}. As DSO provides camera poses for every 3rd frame, we compute metrics using these frames only. Qualitative results are shown in Figure \ref{fig:trajectory_images}. Overall, both ORBSLAM2 and DSO are prone to tracking loss and demonstrate high failure rate. In many cases this problem arises in low-texture scenes, e.g. environments with white walls. But for the sequences where ORBSLAM2 succeeds, it demonstrates very good accuracy. In our experiments DSO showed high scale drift and often lost tracking. Learning-based methods are more robust and accurate.

\subsection{Mapping}

\textbf {Evaluated method.}
We chose Open3D \cite{zhou2018open3d} as a baseline algorithm. Open3D produces voxel maps from sequences of RGB-D frames. Taking color, depth and camera extrinsic and intrinsic matrices Open3D extracts point cloud from each frame, transforms it from camera coordinate system to world coordinate system and adds it to point cloud accumulator. We choose Open3D truncated signed distance function (TSDF) as data accumulator. TSDF speeds up point cloud aggregation and makes it much more uniform. It also allows scene to be represented with adjustable level of detail. We select scalable TSDF as more RAM-intelligent point cloud accumulator with resolution 0.03125 m and 0.25 m as truncation threshold.

At the last step Open3D transforms TSDF back to point cloud and we project it onto ground plane as it described earlier. As result we get a 2D predicted map in grid coordinate system and now are able to compare it with ground truth map.

For trajectory estimation we used Motion Maps method \cite{slinko2019scene}, as it showed the best performance. Example of a map produced by Open3d is shown in Figure \ref{fig:maps_example}. We believe that these results can be further improved by using further map optimizations, e.g. with the use of ICP, pose graph optimization or bundle adjustment. The quantitative results are shown in Table \ref{tab:open3d_results}.

\begingroup
\setlength{\tabcolsep}{2.5pt} 
\begin{figure*}[t]
  \begin{tabular}{ccccc}
  \includegraphics[width=0.19\linewidth]{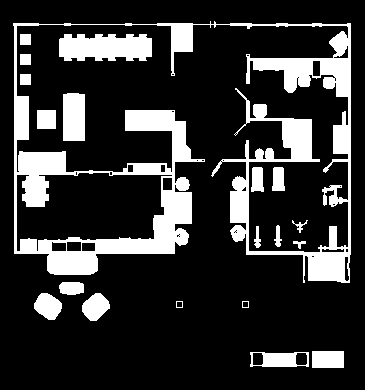} &
  \includegraphics[width=0.19\linewidth]{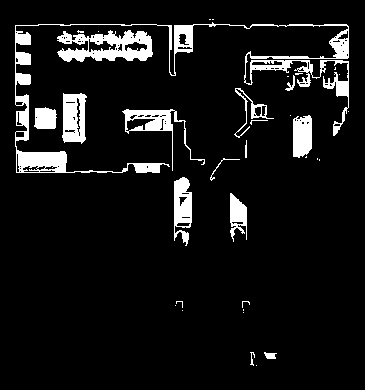} &
  \includegraphics[width=0.19\linewidth]{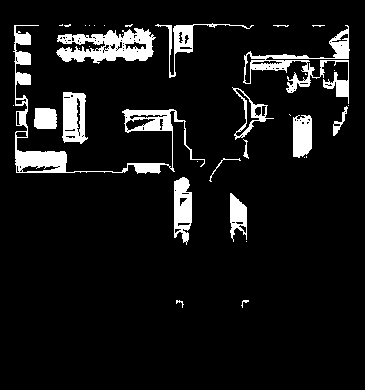} &
  \includegraphics[width=0.19\linewidth]{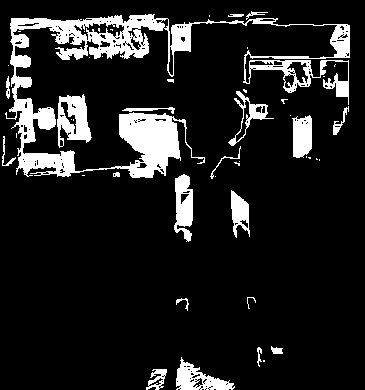} &
  \includegraphics[width=0.19\linewidth]{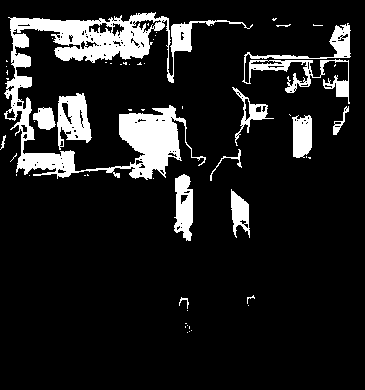} \\
  (a) & (b) & (c) & (d) & (e) 
  \end{tabular}
  \caption{Example of mapping result obtained by Open3D using camera poses from Motion Maps method. (a) - occupancy grid of a 3d scene, (b) - occupancy grid obtained using Open3D with ground truth camera poses and ground truth depth, which we take for ground truth map, (c) - map from ground truth camera poses and noisy depth, (d) - map from camera poses provided by Motion Maps \cite{slinko2019scene} and ground truth depth, (e) - map from poses from Motion Maps and noisy depth. }
    \label{fig:maps_example}
\end{figure*}
\endgroup

\begin{table}[t]
\begin{center} 
\begin{tabular}{l|cc}
\hline
Method & Success rate & Map Score \\
\hline\hline
Open3D (ground truth poses, noisy depth) & 100\% & 87.1\% \\
Open3D (est. poses, noisy depth) & 50\% & 50.7\% \\
\hline
\end{tabular}
\end{center}
\caption{Evaluation results for mapping. We present results for ground truth camera poses and for the camera poses estimated by Motion Maps, which showed highest accuracy in terms of trajectory estimation.}
\label{tab:open3d_results}
\end{table}

\textbf{Details and results.}
To investigate the impact of different sources of errors on the accuracy of mapping we performed the following experiments. To evaluate the impact of depth noise on mapping we run our mapping evaluation pipeline on ground truth depth and on depth with emulated sensor noise. To evaluate the impact of inaccuracies in pose estimation we run experiments with ground truth camera positions/orientations and predicted ones. The results of the evaluation on DISCOMAN dataset are shown in Table \ref{tab:open3d_results}. One can see that inaccurate pose estimation and noisy depth measurements lead to degradation of accuracy.

\subsection{Semantic/panoptic segmentation}

\textbf{Evaluated methods.} We perform experiments for both RGB and RGB-D semantic segmentation methods. For RGB segmentation we have reimplemented the state-of-the-art DeepLabV3+ architecture \cite{chen2018encoder}. To enable RGB-D segmentation we trained the same architectures with added FuseNet-like \cite{hazirbas2016fusenet} branch. For panoptic segmentation we used author's implementation of AdaptIS \cite{sofiiuk2019adaptis}.

\textbf{Details and results.} For semantic segmentation we trained the networks for 16 epochs with SGD momentum=$0.9$, weight decay $10^{-4}$, and linear learning rate scheduler starting with LR=$0.01$. We used ResNet101 \cite{he2016identity} as a backbone and fine-tuned it with one tenth of the learning rate. Crop size was set to $440$. To reduce overfitting we used the following augmentations: random flip, random scale up to $30\%$ of the crop size, random crop, random blur. We trained the models on 2 GPUs with batch size = $8$ in the experiments with RGB images and batch size = $6$ in the experiments with RGB-D images respectively. For panoptic segmentation we used ResNet-50 as a backbone and trained the network for 180 epochs without point proposals and later for 20 more epochs with point proposals. We present the evaluation results in Table \ref{tab:panoptic_results}.

\begingroup
\setlength{\tabcolsep}{1pt} 
\renewcommand{\arraystretch}{0.8} 
\begin{figure}
  \begin{tabular}{ccc}
    \includegraphics[width=0.323\linewidth]{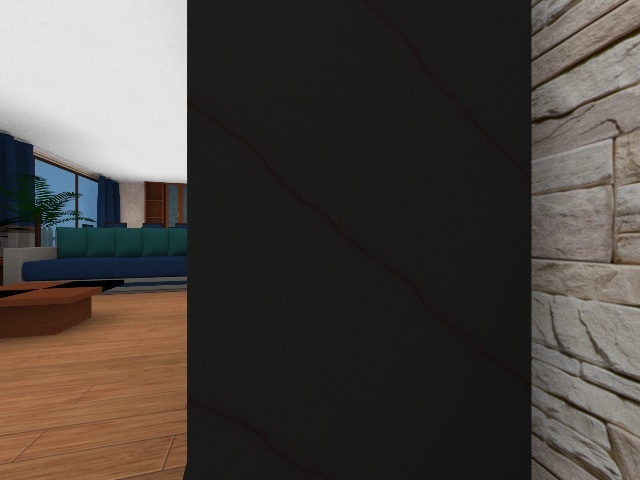} &
    \includegraphics[width=0.323\linewidth]{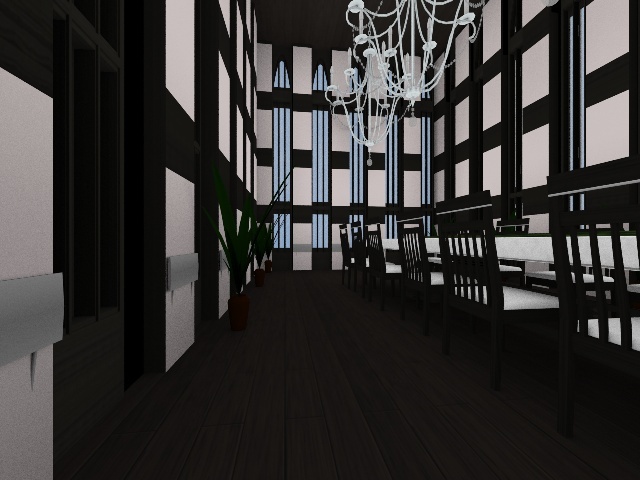} &
    \includegraphics[width=0.323\linewidth]{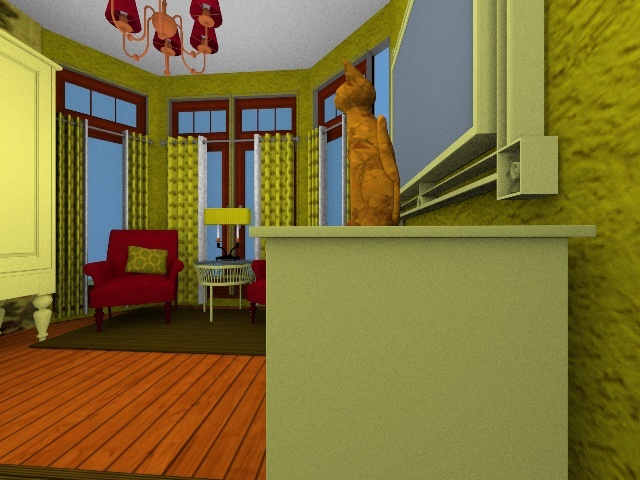} \\
    \includegraphics[width=0.323\linewidth]{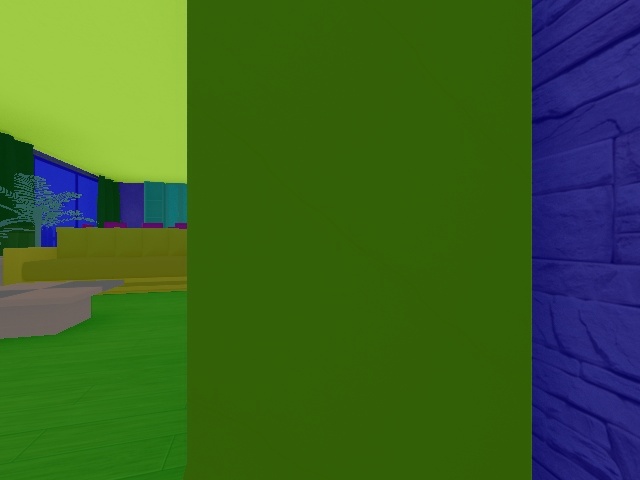} &
    \includegraphics[width=0.323\linewidth]{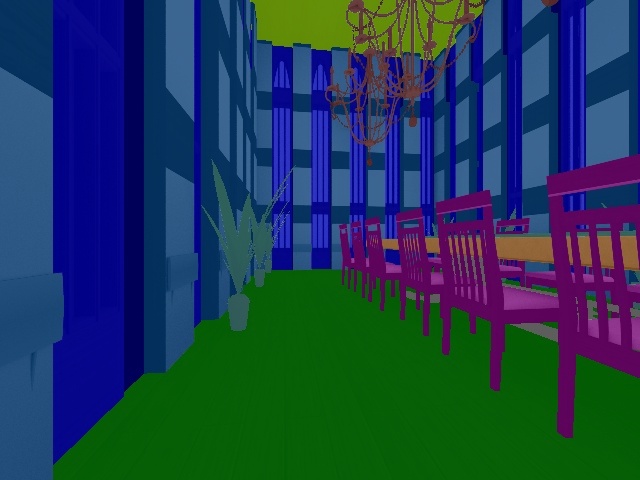} &
    \includegraphics[width=0.323\linewidth]{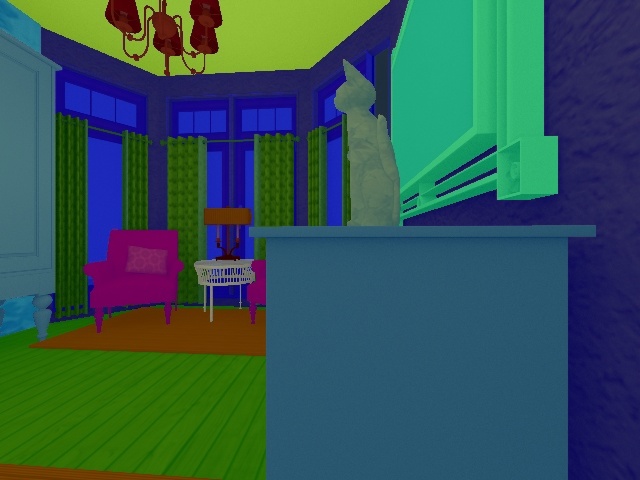} \\
    \includegraphics[width=0.323\linewidth]{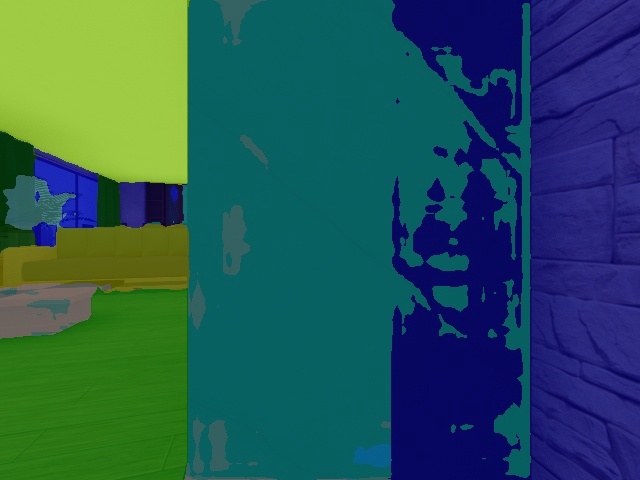} &
    \includegraphics[width=0.323\linewidth]{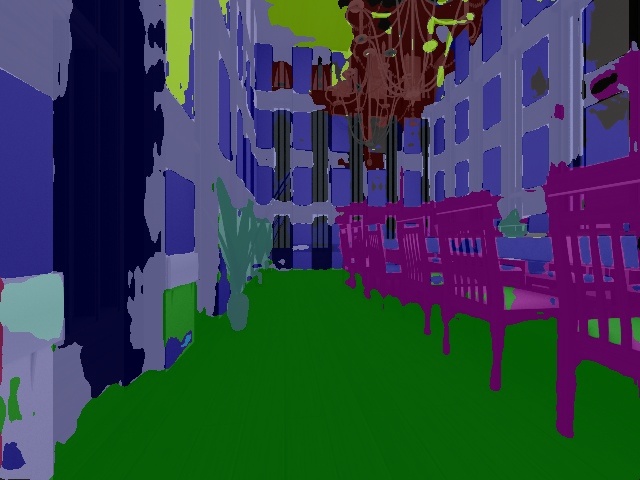} &
    \includegraphics[width=0.323\linewidth]{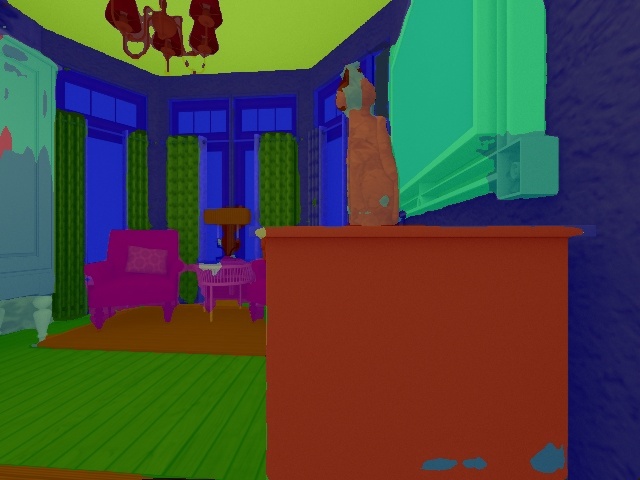} \\
    \includegraphics[width=0.323\linewidth]{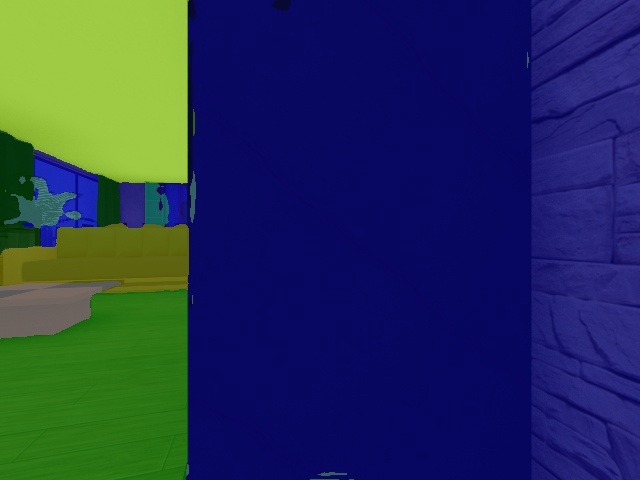} &
    \includegraphics[width=0.32\linewidth]{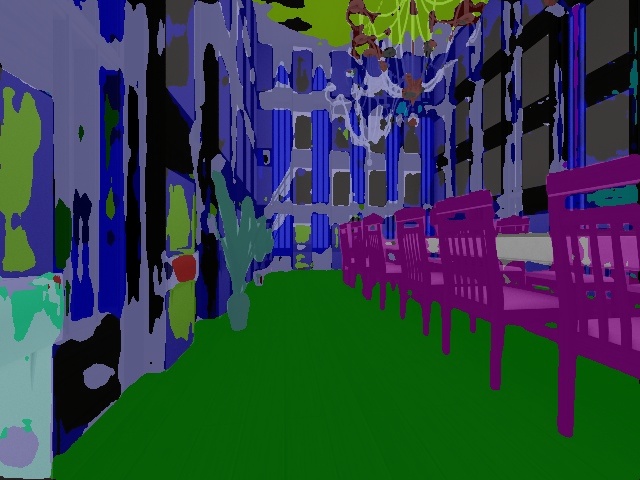} &
    \includegraphics[width=0.32\linewidth]{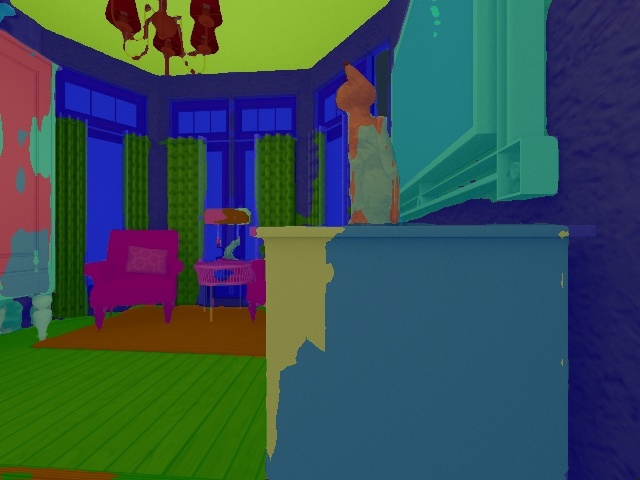} 
  \end{tabular}
  \caption{Failure cases for semantic segmentation. First row -- input image, second row --- ground truth semantic labelling, third row --- result of DeepLabV3+ RGB segmentation, fourth row --- result of DeepLabV3+ RGB-D segmentation. One can see that in some cases adding depth information helps to deal with ambiguities, but overall the effect of using depth for semantic segmentation is not dramatic.}
  \label{fig:segmentation}
\end{figure}
\endgroup

\begin{table}
\begin{center}
\begin{tabular}{l|c|c}
\hline
Method & mIoU & pixel accuracy \\
\hline\hline
DeepLabV3+ (RGB only) & 77.41\% & 95.73\% \\
DeepLabV3+ with FuseNet (RGB-D) & 79.88\% & 96.11\% \\
\hline
\end{tabular}
\end{center}
\caption{Evaluation results for DeepLabV3+ \cite{chen2018encoder} on DISCOMAN dataset. For RGB-D segmentation we added FuseNet-like branch \cite{hazirbas2016fusenet} to DeepLabV3+ architecture.}
\label{tab:segmentation_results}
\end{table}

\begin{table}
\begin{center}
\begin{tabular}{l|c|c|c}
\hline
 & PQ & SQ & RQ \\
\hline\hline
All & 50.22 & 83.27 & 57.18 \\
Things & 46.61 & 81.87 & 53.41 \\
Stuff & 62.59 & 88.05 & 70.10 \\
\hline
\end{tabular}
\end{center}
\caption{Evaluation results for AdaptIS \cite{sofiiuk2019adaptis} on DISCOMAN dataset.}
\label{tab:panoptic_results}
\end{table}

The results of our experiments are shown in Table \ref{tab:segmentation_results}. Qualitative results and failure cases are shown in Figure \ref{fig:segmentation}. One can notice that adding information about depth leads to slightly improved accuracy of semantic segmentation.

\section{Conclusion} \label{conclusion}
\label{sec:conclusion}
We have presented a new dataset and benchmark suite for training and evaluation of semantic SLAM models. This is the first large-scale dataset that provides ground truth annotation for environment maps in the form of occupancy grids. We present benchmarking results for RGB/RGB-D SLAM, mapping and semantic/panoptic segmentation methods across conventional metrics to establish baselines for further research.

\section*{Acknowledgments}


\bibliographystyle{ieeetr}
\bibliography{references}

\end{document}